\pdfoutput=1

\documentclass[11pt]{article}

\usepackage[]{emnlp2022}

\usepackage{times}
\usepackage{latexsym}
\usepackage{adjustbox}
\usepackage{booktabs}

\usepackage{array, booktabs, multirow}
\newcolumntype{C}[1]{>{\centering\arraybackslash}p{#1}}

\usepackage[T1]{fontenc}

\usepackage[utf8]{inputenc}

\usepackage{microtype}

\usepackage{subcaption}
\usepackage{inconsolata}
\usepackage{enumitem}
\usepackage{wrapfig}
\usepackage{graphicx}

\usepackage[normalem]{ulem}

\title{BIG-C: a Multimodal Multi-Purpose Dataset for Bemba}

\author{Claytone Sikasote$^1$, Eunice Mukonde$^2$, Md Mahfuz Ibn Alam$^3$, Antonios Anastasopoulos$^3$ \\
$^1$Department of Computer Science, University of Zambia, Zambia\\
$^2$Department of Literature and Languages, University of Zambia, Zambia\\
$^3$Department of Computer Science, George Mason University, USA \\
         \texttt{claytone.sikasote@cs.unza.zm}, \texttt{antonis@gmu.edu}\\
}

\begin{document}
\maketitle
\begin{abstract}
We present BIG-C (Bemba Image Grounded Conversations), a large multimodal dataset for Bemba.  While Bemba is the most populous language of Zambia, it exhibits a dearth of resources which render the  development of language technologies or language processing research almost impossible. The dataset is comprised of multi-turn dialogues between Bemba speakers based on images, transcribed and translated into English.  There are more than 92,000 utterances/sentences, amounting to more than 180 hours of audio data with corresponding transcriptions and English translations. We also provide baselines on speech recognition (ASR), machine translation (MT) and speech translation (ST) tasks, and sketch out other potential future multimodal uses of our dataset. We hope that by making the dataset available to the research community,\footnote{All data and code are publicly available: \url{https://github.com/csikasote/bigc}.} this work will foster research and encourage collaboration across the language, speech, and vision communities especially for languages outside the "traditionally" used high-resourced ones.
\end{abstract}

\section{Introduction}

\begin{figure}[h!]
\centering
\tiny
\begin{tabular}{>{\raggedright\arraybackslash}p{.2\textwidth}>{\raggedleft\arraybackslash}p{.2\textwidth}}
    \multicolumn{2}{c}{\includegraphics[width=.35\textwidth]{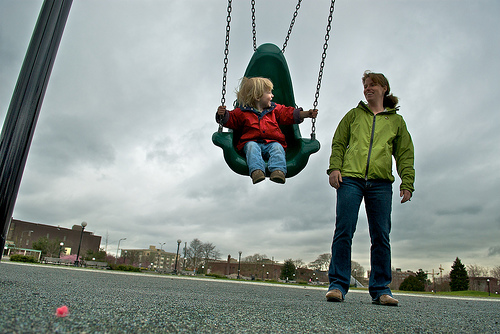}} \\
    \includegraphics[width=.6cm]{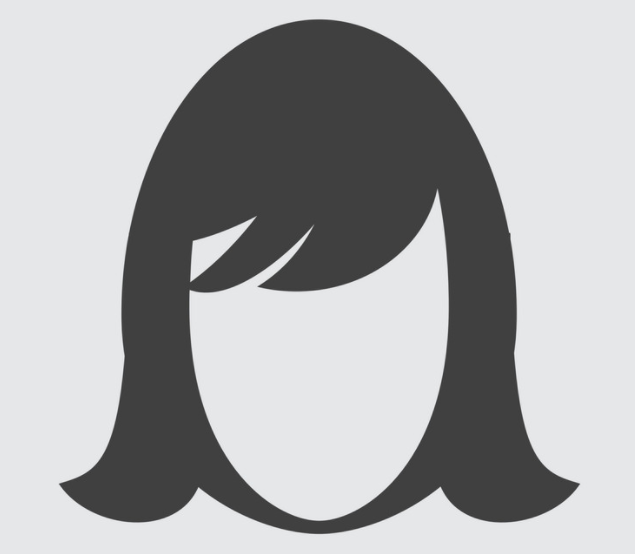} \includegraphics[width=2cm]{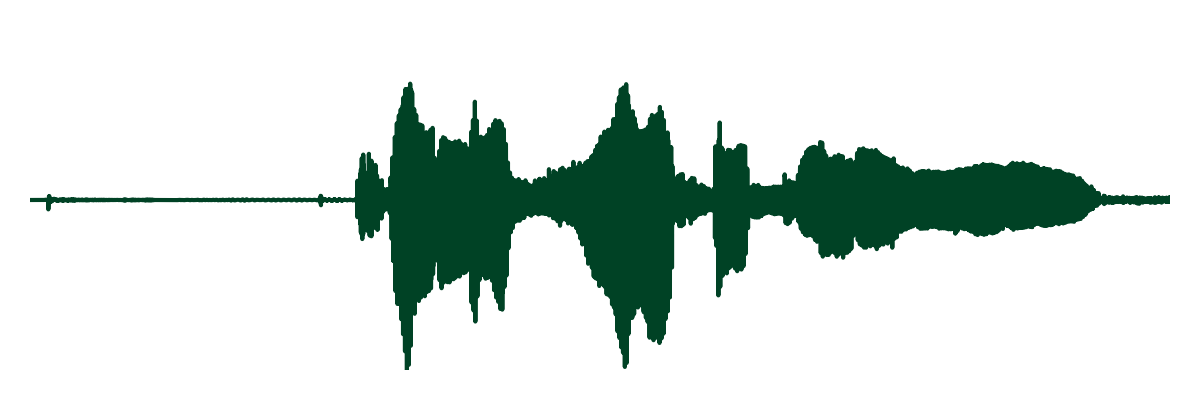} &  \\[-2em]
    & \includegraphics[width=2cm]{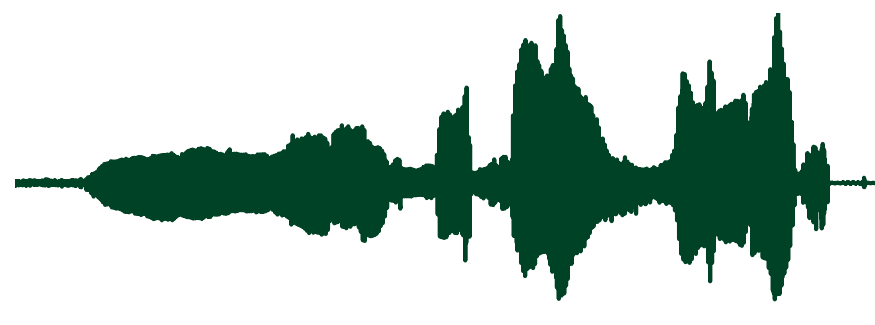} \includegraphics[width=.6cm]{images/face.png} \\[-2em]
        
    Umwana alepeluka pa cimpelwa elo ba nyina pa mbali balemutamba & \\
    \textcolor{gray}{\textit{A child is on a swing chair swinging while the mother watches on the side}} & \\[-3em]
     & Akamwana kali akasansamuka saana,kalelosha ba nyina nakasekelela. \\
     & \textcolor{gray}{\textit{The child is very excited for swinging, looking at its mother with a smile.}} \\[-1em]
     Cipalile pali efyo balekeba ba nyina pantu bonse bali abasekelea &  \\
     \textcolor{gray}{\textit{It seems like she is telling the child something as they both have smiled.}} & \\[-2em]
     & Limbi kaleumfwafye bwino filyafine kalepeluka nabanyina balekasunka. \\
     & \textcolor{gray}{\textit{It may be excited for swinging even as the mother swings it.}}\\[-2em]
     Awe cine abana balatemwa banyina, insansa muli uyu umwana shafulisha & \\
     \textcolor{gray}{\textit{Indeed children love their mothers, the child is just too excited.}} \\
\end{tabular}
\caption{Example of the data included in BIG-C. The grounding image (top) and the ensuing Bemba dialogue transcribed and translated in English.}
\label{fig:example}
\vspace{-1em}
\end{figure}

\begin{table*}[t]
\centering
\small
\begin{tabular}{p{6.5cm}lllp{2cm}l}
\toprule
& \textbf{Images} & \textbf{Text} & \textbf{Audio} & \\
\textbf{Dataset} & \textbf{(\#unique)} & \textbf{(turns)} & \textbf{(hours)} & \textbf{Languages(s)} & \textbf{Parallel}\\
\midrule
\multicolumn{5}{@{ }l}{\textbf{Task: Image Captioning}}\\
\verb|MSCOCO|~\footnotesize{\cite{Lin2015}} & 330K & 1.5M & - & Eng & NA\\
\verb|Flickr8K Audio |~\footnotesize{\cite{Harwath2016}} & 8K & 40K & 65 & Eng  & NA\\
\verb|Flickr30K|~\footnotesize{\cite{plummer2015flickr30k}} & 30K & 158K & - & Eng & NA\\
\verb|Pascal Sentences |~\footnotesize{\cite{funaki-nakayama-2015-image}} & 1K & 10K & - & Eng, Jap & Partial\\
\verb|IAPR TC-12 |~\footnotesize{\cite{grubinger2006iapr}} & 1K & 10K & - & Eng, Deu, Spa & No\\
\verb|Multi30K|~\footnotesize{\cite{Elliott2016,elliott-etal-2017-findings,barrault-etal-2018-findings}} & 30K & 155K & - & Eng, Deu, Fra, Ces  & Yes\\
\verb|WIT |~\footnotesize{\cite{Srinivasan2021}} & 11.5M & 37.6M & - & 108 langs & Partial\\ 

\verb|HaVG |~\footnotesize{\cite{abdulmumin-etal-2022-hausa}} & 30K & 30K & - & Eng, Hau & Yes\\
\verb|BAN-Cap |~\footnotesize{\cite{khan-etal-2022-ban}} & 8K & 40K & - & Eng, Ben & Yes\\
\verb|Bloom Library|~\footnotesize{\cite{leong-etal-2022-bloom}} & 90K & 110K & 428 & 363 langs & NA\\
\multicolumn{5}{@{ }l}{\textbf{Task: Dialogues over Images}}\\
\verb|IGC|~\footnotesize{\cite{mostafazadeh-etal-2017-image}} & 4.2K & 25K & - & Eng  & NA\\
\verb|Image-Chat|~\footnotesize{\cite{shuster-etal-2020-image}} &  202K &  202k & -- & Eng & NA\\
\midrule
\verb||\textbf{BIG-C} & 16K & 90K & 185 & Bem, Eng  & Yes\\
\bottomrule
\end{tabular}
\caption{BIG-C and related datasets. BIG-C is the only \textit{multi-purpose} dataset in an under-served language.} 
\label{tab:related-dataset}
\vspace{-1em}
\end{table*}

The Bemba language, spoken by over 10 million people in Zambia and other parts of Africa, is a rich and vibrant language with a unique cultural heritage. However, despite its significance, Bemba is a dramatically under-resourced language, lacking in high-quality language data and resources for natural language processing (NLP) experiments and for the development of language technologies. With this work, we address this issue by creating a new multimodal dataset for Bemba. Our goal is to improve the accuracy and effectiveness of NLP systems for speakers of Bemba and support research in this under-served language.

While most datasets are constructed with a specific task in mind and tailored to its characteristics, we aim to provide a path towards building \textit{multi-purpose} datasets. Under a limited budget, we hypothesize that the ideal scenario is to create datasets that can be useful for developing multiple language technologies for both practical applications and also facilitate cutting-edge NLP research on many dimensions. Our hope is that such datasets will aid in bridging the ever-widening language divide both in terms of data availability~\cite{joshi-etal-2020-state} and NLP research~\cite{blasi-etal-2022-systematic}, and make language technologies more accessible for speakers of Bemba.

In this work, we present our methodology and results of creating a new multimodal dataset for Bemba, and demonstrate the potential of this dataset to develop NLP systems and support NLP research.
Our dataset will fill multiple roles: enable development of fundamental tools such as speech recognition, speech and text translation systems for Bemba; serve as a benchmark for academic and industry research even as NLP for low-resource and under-represented African languages gets developed; facilitate research in language grounding and multimodal model development, or building context-based dialogue agents, among other possible use cases. To our knowledge this is the first such dataset of its kind for any Zambian and possibly African language. We hope that it will provide an example of how to create a \textit{multi-purpose} dataset in an under-served language to facilitate its coverage by multiple technologies.

The rest of the paper is structured as follows: in Section 2,  we briefly introduce the Bemba language discussing any currently available resources.  In Section 3, we summarise work related to multimodal tasks and existing datasets. In Section 4, we provide a description of the BIG-C dataset and the methodology used, and in Section 5, we provide baseline experiments for some NLP tasks. 

\section{The Bemba Language}

Bemba, also known as \textit{IciBemba} or \textit{Cibemba}, is a Bantu language native to Luapula, Muchinga and Northern provinces of Zambia. It is also spoken in other urban parts of the country like Copperbelt, Central and Lusaka provinces. It is estimated that Bemba is spoken by over 30\% of the population of Zambia as either the first or second language, making it the language with the most speakers in the country~\cite{Kapambwe2018}. A map of Bemba usage in Zambia is provided in Appendix Figure~\ref{fig:map}.

The Bemba language has a number of dialects and the main varieties are: Standard Bemba also Central Bemba, Aushi, Bisa, Chishinga, Lamba, Lala, Luunda, Ngumbo, Swaka, Tabwa and Unga. These dialects show minor differences in phonology, morphology and vocabulary\cite{Spitulnik2001,Spitulnik2014}. In this work, we focus on the Standard Bemba dialect, i.e., the one spoken in urban centers around the country. 

\paragraph{Datasets for Bemba}

For ASR, to the best of our knowledge, there is only a single dataset publicly available for Bemba, BembaSpeech~\cite{sikasote-anastasopoulos-2022-bembaspeech}. It contains 24 hours of read-styled speech data recorded from text mainly sourced from various source but mainly literature books. The low resource nature of the BembaSpeech~\cite{sikasote-anastasopoulos-2022-bembaspeech} dataset makes it difficult to build usable ASR system for Bemba.  For machine translation (text-to-text), there is not a single dedicated dataset for Bemba. However, there exist some parallel text-to-text data in multilingual datasets such as JW300~\cite{Agic2020} and in evaluation benchmarks such as NTREX-128~\cite{federmann-etal-2022-ntrex} and FLORES-200~\cite{nllb2022}. The text in the JW300~\cite{Agic2020} is mostly religious as it is derived from the Bible text. For speech translation (speech-to-text; ST), to our knowledge, no prior work or Bemba dataset exists. This essentially renders it impossible to build a ST system where Bemba is a source or target language. The same is true for multimodal and dialogue datasets: there is no multimodal or dialogue-related dataset for any Zambian language that would enable development of multimodal systems. Our work aims to fill these gaps. 

\section{Related Work}
\label{sec:related-work}
In the recent years, NLP, speech processing (SP) and computer vision (CV) fields have rapidly advanced, with computational models' performance achieving new heights on a wide range of downstream tasks. This, to some degree, can be attributed to factors such as the emergence of pre-trained models leveraging self-supervised learning, the availability of large-scale datasets, and increased large-scale computational infrastructure~\cite{Hirschberg2015}. In NLP, language models like BERT~\cite{devlin-etal-2019-bert}, T5~\cite{2020t5}, GPT3~\cite{Brown2020} and XLM-R~\cite{Conneau2020}, pretrained on massive text datasets such as C4~\cite{2020t5}, mC4~\cite{Xue2021} and BooksCorpus~\cite{Zhu_2015_ICCV} among others, have lead to significant performance improvements on several language understanding and generation downstream tasks. Likewise, for speech processing, the unsupervised pretraining of models like wav2vec2.0~\cite{Baevski2020} or XLS-R~\cite{babu2021xlsr} -- having been pretrained on publicly available speech datasets such as VoxPopuli~\cite{wang-etal-2021-voxpopuli}, MLS~\cite{Pratap2020MLSAL}, Commonvoice~\cite{commonvoice:2020}, BABEL~\cite{BABEL:CVPR:2021} among others, have led to advances on speech downstream tasks like ASR~\cite{babu2021xlsr} and ST. In computer vision, deep learning models like DeepCNN~\cite{Simonyan2015, He2016} have become the \textit{de facto} solution for standard vision problems like object recognition~\cite{He2016}, image classification~\cite{Krizhevsky2017}, or semantic segmentation~\cite{Shelhamer2017}.

Since these neural models are conceptually (and architecturally) quite similar they have also enabled the integration of multiple modalities, with models such as ViLBERT~\cite{Lu2019}, UNITER~\cite{chen2020uniter}, Unicoder-VL~\cite{huang2019unicoder} able to jointly model the relationship between text and image modalities resulting into breakthroughs across a myriad of tasks such as image-text retrieval/search~\cite{Frome2013, Liu2020}, image or video captioning~\cite{Biten2019}, and vision-question answering ~\cite[VQA;][]{Agrawal2017, Nam2017}. A crucial necessary component for all of the above, of course, is the availability of relevant datasets. Below we discuss works that go beyond the collection of raw datasets that are used for self-supervised learning.

\paragraph{Dialogue}
In the recent past, a lot of work has been focused on dialogue datasets. On one hand there exist goal-oriented dialogue datasets, such as the case of the Ubuntu dialogue corpus~~\cite{lowe2015ubuntu}, the largest corpus of dialogues (almost 1 million mainly 3-turn dialogues in English) for the specific topic of troubleshooting Ubuntu problems. On the other hand, open ended conversations, such as those on the CALLHOME/CALLFRIEND~\cite{Canavan1997} or Fisher corpora~\cite{cieri-etal-2004-fisher}, often leads to un-interesting conversations. Grounding the dialogue to event-centric images and potentially a specific scenario  constrains the topic of conversation to event-rich and contentful utterances.

\paragraph{Multimodality} Multimodal works combining visual and language information typically focus on image captioning and visual question answering~\cite{antol2015vqa}. For example, the IAPR TC-12 dataset~\cite{grubinger2006iapr} provides images with titles and descriptions (mostly in English, German, and Spanish), as do commonly used datasets like MSCOCO~\cite{Lin2015} and Flickr30K~\cite{plummer2015flickr30k}. Flickr8K Audio~\cite{Harwath2016} extended a subset of the Flickr images with audio, by crowdsourcing readings of the English captions, while Multi30K~\cite{Elliott2016} further extended Flickr30K with German translations and annotations. Wikipedia-based Image Text (WIT) Dataset~\cite{Srinivasan2021} provided large multilingual coverage (108 languages) based on 11.5M images and captions from Wikipedia. More recent, Hausa Visual Genome \cite[HaVG;][]{abdulmumin-etal-2022-hausa} provided over 30K parallel descriptions in English and Hausa of images from the Hindi Visual Genome \cite[HVG;][]{Parida2019HvG}. The dataset was created by automatically translating the English descriptions of the images in the HVG to Hausa using Google Translate\footnote{\url{https://translate.google.com/}} and post-edited by crowd-sourced Hausa volunteers.  Similarly, BAN-Cap~\cite{khan-etal-2022-ban} provides over 40K human-annotated parallel English-Bangla image description pairs based on 8,091 images from Flickr8K~\cite{Harwath2016}. Lastly, the Bloom Library~\cite{leong-etal-2022-bloom} provides a set of multilingual datasets for language modeling, image captioning and visual-story telling tasks containing more than 110K image captions for over 90K images in 351 languages. It also provides a speech dataset with 428 hours of speech data for speech synthesis/recognition tasks covering 56 languages.

Beyond captioning tasks, the dialog component was first explored by~\citet{das2017visual}, who extended the VQA scenario collecting sequential questions grounded on images. \citet{mostafazadeh-etal-2017-image} went beyond goal-oriented dialogue to collect image-grounded conversations (contrasting this to open-ended dialogue research). More recently, the Image-Chat dataset \cite{shuster-etal-2020-image} collected open-ended  conversations grounded in images with a focus on engagement, by assigning desired style traits to the speaker.

\paragraph{Discussion} 
There are notable limitations with most publicly available multimodal datasets. To make comparisons easy, we outline most relevant works in Table~\ref{tab:related-dataset}. While the list shown there is non-exhaustive, these limitations can be grouped in terms of language coverage, modality composition, tasks supported i.e., single-purpose or multi-purpose tasks. To give more context to this categorization:
\begin{itemize}[noitemsep,nolistsep]
    \item In terms of languages, they cover only a handful of high-resourced languages like English.
    \item In terms of modality composition, the majority only contain image and text modalities, ignoring the audio component.
    \item With regards to tasks, the majority are meant for a single-purpose task such as image captioning.\footnote{An exception to this is the Bloom Library~\cite{leong-etal-2022-bloom}. But note that it lacks representation of any Zambian language among the covered languages.}
\end{itemize}

In contrast, our work presents a \textit{multimodal} but also \textit{multi-purpose} dataset for Bemba. Our aim is for BIG-C to be the first-of-its-kind dataset for an under-served language that can \textit{simultaneously} serve as:
\begin{itemize}[noitemsep,nolistsep]
    \item a monolingual dataset for Bemba e.g., to be used for training language models on this under-served language;
    \item a parallel dataset to allow for building and evaluating machine translation solutions;
    \item an image captioning dataset with image descriptions in Bemba;
    \item an image-grounded dialogue dataset;
    \item a benchmark for any combination between the above modalities e.g., one could use our dataset to evaluate image-grounded dialogue translation systems.
\end{itemize}
We achieve this through careful instructions and data collection practices, outlined in Section \S\ref{sec:dataset}.

\section{Dataset Description}
\label{sec:dataset}
\paragraph{Description}
The dataset consists of a parallel corpus of speech and transcriptions of image-grounded dialogues between Bemba speakers and their corresponding English translations.  It contains 92,117 spoken utterances (complete and incomplete dialogues), amounting to 187 hours of speech data grounded on 16,229 unique images. There are 16,697 complete 5-turn unique dialogues grounded on 14,551 unique images. Of the total 16,697 complete dialogues, 2,146 are unique dialogues grounded on duplicated images, each recorded by unique pairs of speakers. A second set of dialogues is comprised of 2,314 incomplete dialogues missing one or more utterances as a result of the pre-processing step that involved removing all audio files that are silent and corrupted. The sum of utterances that make up the incomplete dialogues is 8,632 of the total 92,117 utterances. All audio files are encoded in Waveform Audio File format (WAVE) with  a single track (mono) and sample rate of 16kHz. In Table~\ref{tab:dataset-statistics}, we provide basic dataset statistics.

\begin{table}[t]
\centering
\begin{tabular}{p{5.5cm}c}
\toprule
\textbf{Description} & \textbf{Count} \\
\midrule
\multicolumn{2}{@{}l}{\textbf{Data}} \\
\#  unique images & 16,229  \\
\#  hours transcribed and translated & 187 \\ 
\#  complete dialogues & 16,697  \\
\#  "incomplete" dialogues & 2,314 \\
\#  sentences/complete dialogue & 5  \\
\#  spoken utterances & 92,117  \\
\#  English translations & 92,117 \\ 
\#  Bemba tokens & 870K \\
\#  English tokens & 1.1M \\
\multicolumn{2}{@{}l}{\textbf{Metadata}} \\
\#  speakers  & 86 \\
\#  transcribers & 93 \\
\#  translators & 114 \\
\#  validators & 15 \\
\bottomrule
\end{tabular}
\begin{tabular}{lc}
\hline
\end{tabular}
\caption{BIG-C: Basic Dataset Statistics.} 
\label{tab:dataset-statistics}
\end{table}

\paragraph{Source of images}
 We randomly selected images from the Flickr30K~\cite{plummer2015flickr30k} dataset, a publicly available multimodal dataset for vision and language that has become a standard benchmark for sentence-based image descriptions.

\paragraph{Speakers} To record conversations, we recruited 86 speakers of the Bemba language; 60\% male and 40\% female, based on their competency to speak, read and write the language. Based on the metadata information supplied by participants, we summarise the characteristics of our speakers as follows:

\begin{itemize}[noitemsep,nolistsep,leftmargin=*]
    \item \textbf{Age:} the majority of the speakers (98\%) were youth whose age falls between 20 and 35 years old with the 2\% being over 35 years old.
    \item \textbf{Education:} all speakers had some form of secondary education; 90\% of the participant were either pursuing or recently graduated with a college/university degree; and the rest 8\% had only completed high school. 
    \item \textbf{Language(s):} all speakers were bilingual; with 90\% indicating Bemba as their first language and Nyanja as the majority non-English second language. 
    \item \textbf{Regions:} in terms of regional representations, over 90\% of the speakers were drawn from Lusaka, Central, and Copperbelt regions; with small representations from Muchinga and Northen provinces. This in effect indicates that the dataset is composed of the current 'urban' Bemba variety.
    \item \textbf{Racial diversity:} the composition of our participants lacks racial diversity, as all speakers are identified as black.
\end{itemize}

\begin{table*}[t]
\centering
\begin{tabular}{lcccccc}
\toprule
\multicolumn{7}{r}{\textbf{No. of speaker voices}} \\
\cmidrule(r){5-7}
\textbf{Split} & \textbf{Images} &  \textbf{utterances} & \textbf{hours} & \textbf{Male} & \textbf{Female} & \textbf{Unspecified} \\
\midrule
Train & 14,599 & 82,375 & 167 & 43,959 & 38,338 & 78 \\
Valid & 492 & 2,782 & 5 & 1,491 & 1,289 & 2 \\
Test & 501 & 2,779 & 5  & 1,457 & 1,318 & 4 \\
Held  & 637 & 4,181 & 8  & 2,105 & 2,072 & 4 \\
\midrule
\textbf{Total} & 16,229 & 92,117 & 185  & 49,012 & 43,017 & 88 \\
\bottomrule
\end{tabular}
\caption{\label{tab:dataset-split-hours}Summary details of the splits of the dataset.}
\vspace{-1em}
\end{table*}

\paragraph{Recording} The speakers were randomly paired with gender-balancing in mind.  Each pair was allocated 250 images to create 5 sentence-turn conversation per image for each recording session. There was no restriction to what each pair would converse about on an image. The participants were encouraged to be creative. However, the conversation starter (speaker 1) was instructed to first describe the image, so as to give context to the conversation (and essentially provide data for the image captioning component of our dataset). We provide the sample instructions that were given to the annotators in Appendix~\ref{sec:training-instruction}. All recordings were conducted in minimally controlled conditions. The pairs recorded as per their comfort, we therefore expect that some spoken utterances have background noise. All participants used the LIG-AIKUMA~\cite{Gauthier2016} mobile application, using the `elicitation by image' mode to record spoken utterances.

\paragraph{Transcribers} To transcribe the audio data generated from the image-grounded conversations, we recruited 93 participants, who in their majority were students of the University of Zambia. 
All were competent Bemba speakers and writers. As shown in Table~\ref{tab:dataset-statistics}, 92,117 spoken utterances were transcribed representing 187 hours of Bemba speech data.

\paragraph{Translators} To translate a subset of the transcriptions to English, we recruited 115 participants with experience in translating Bemba text to English or vice versa. Public education in Zambia is conducted in English, hence we are confident in a minimum translation quality.

\paragraph{Splitting} We have split the dataset into training, validation and testing sets following the original splits in the Flickr30K~\cite{plummer2015flickr30k} dataset according to the images. See Table~\ref{tab:dataset-split-hours} for more details.

\paragraph{Data quality} 
Several measures were set up during the data collection process to ensure quality submissions from project participants; speakers, transcribers and translators. First, at recruitment stage for audio recording, we considered only competent Bemba speakers with ability to speak, read and write in Bemba. All the speakers underwent a training exercise to make sure they understood and followed instructions of how to go about the task of creating and recording multi-turn conversations using the Lig-Aikuma~\cite{Gauthier2016} mobile application. For the transcriptions, we retained good number of the speakers - over 50\% to also participate in transcribing the audio files at transcribing stage. In addition, we recruited validators, who together with the authors of this study checked and verified manually every submission made by the participants at every stage of the process. All audio files that were deemed to be of low quality i.e., silent, corrupted and inaudible due to background noise, were removed as part of data pre-processing at the quality assurance and validation stage.

Last, during the translation stage, besides the ability to speak, read and write, we recruited participant who had experience with translating Bemba text to English as translators. Most of the participants had prior experience as professional or volunteer translators.

\paragraph{Availability} The dataset is made available to the research community licensed under the Creative Commons BY-NC-ND 4.0 license and can be accessed at our Github repository.\footnote{\url{https://github.com/csikasote/bigc}} We do plan to keep a small held-out portion unpublished, to be used in future shared tasks or as part of leaderboards that require \textit{hidden} test sets to ensure a fair measure of task progress.

\section{Baseline Experiments}
In this section, we detail some baseline experiments carried out to demonstrate the potential of the dataset. We provide unimodal baselines using the train-validation-test splits in Table~\ref{tab:dataset-split-hours} on the following tasks: ASR for Bemba, MT and ST of Bemba utterances to English text.

\paragraph{Data preprocessing} 
For ASR and ST, similar to~\citet{wang-etal-2020-covost}, all text i.e., transcriptions and translations, we lower the cases and remove punctuation except for apostrophes, and build 1K unigram character vocabularies with 100\% coverage of all the characters using SentencePiece~\cite{kudo-richardson-2018-sentencepiece} without pre-tokenization. We extract 80-dimensional log-mel scale filterbank features from Bemba utterances using a 25ms window size and 10ms window shift using torchaudio.\footnote{https://github.com/pytorchaudio} The features are normalized to 0 mean and 1.0 standard deviation. All models are trained without an auxillary language model.

\paragraph{Model Architecture}  We use the small Transformer~\cite{Vaswani2017} base architecture with 71 M parameters, \texttt{s2t\_transformer\_s}, having 12-layers encoder, 6-layers decoder, and hidden dimension D=256 to train end-to-end (E2E) ASR and ST models using FAIRSEQ S2T Toolkit~\cite{ott2019fairseq, wang2020fairseqs2t}. Models are trained on a single NVIDIA Tesla P100 GPU using the Google Colab+ platform.


\subsection{Automatic Speech Recognition}
For the ASR baseline model for Bemba, we trained the model for 500 epochs using the Adam optimiser~\cite{Kingma2015} with 10K warm up steps. The model is optimised to minimise the \texttt{label\_smooth\_cross\_entropy} criterion function using the learning rate coefficient of \texttt{2e-3}. For decoding, we use the beam search algorithm with a beam size of 5. We use the average of the last 5 checkpoints for evaluation. In Table~\ref{tab:results}, we report the model performance on the Test set using word error rate (WER) metric. 

\subsection{Speech Translation}
For speech to text translation of Bemba spoken utterances to English text, we use the same model architecture as ASR. The model is trained with same configuration as the ASR model except we use the learning rate coefficient of \texttt{3e-4}. Similarly, we use the beam search algorithm with beam size of 5 for decoding. We use the best checkpoint to evaluate the model on the test set. We report the detokenised case-sensitive BLEU~\cite{Papineni2002} using sacreBLEU~\cite{Post2018} in Table~\ref{tab:results}. 

\paragraph{Evaluation} We use beam search with a beam size of 5 for decoding. We use the average of the last 5 checkpoints to evaluate both ASR and the best checkpoint saved for ST model. We report the results in Table~\ref{tab:results}. For ST, we report detokenised case-sensitive BLEU~\cite{Papineni2002} using sacreBLEU~\cite{Post2018} and word error rate (WER) for ASR. 

\paragraph{Results discussion} For both ASR and ST, we consider the results obtained decent for the size of our dataset and the basic training configurations of our baseline models, which are without auxillary models, and mostly relied on default settings in the FAIRSEQ S2T implementation. We believe the results can be improved upon, and we leave the full exploration of the best configurations to future work. We encourage the community to improve upon these baselines, for instance, by exploring cross-lingual transfer learning by leveraging large scale multilingual pretrained models like XLS-R~\cite{babu2021xlsr} and Whisper~\cite{radford2022whisper}.

\begin{table}[t]
\centering
\begin{tabular}{ll}
\toprule
\textbf{Task} & \textbf{Metric: Value} \\
\midrule
Speech Recognition & WER ($\downarrow$): 32.7 \\
\midrule
Speech Translation & BLEU ($\uparrow$): 17.9 \\
\bottomrule
\end{tabular}
\caption{Baseline results (test set) on speech-based tasks. For ST, we report detokenised case-sensitive BLEU~\cite{Papineni2002} using sacreBLEU~\cite{Post2018} and word error rate (WER) for ASR.} 
\label{tab:results}
\end{table}

\begin{table*}[t]
    \centering
    \begin{tabular}{c|cc|cc}
    \toprule
      & \multicolumn{2}{c|}{BIG-C} & \multicolumn{2}{c}{FLORES-200} \\
        \textbf{Model} & \textbf{\textbf{eng$\rightarrow$bem}} & \textbf{\textbf{bem$\rightarrow$eng}} & \textbf{\textbf{eng$\rightarrow$bem}} & \textbf{\textbf{bem$\rightarrow$eng}}  \\
    \midrule
        DeltaLM FT &  17.9 & 27.5 & 3.5 & 4.3\\ 
        Phylogeny FT & 16.5 & \textbf{28.9} & \textbf{6.0} & \textbf{18.0} \\
    \bottomrule
    \end{tabular}
    \caption{Baseline text translation results. The phylogeny-based model benefits from parameter sharing across all the other Bantu languages.}
    \label{tab:results-mt}
    \vspace{-1em}
\end{table*}

\subsection{Machine (Text) Translation}

For Machine Translation we rely on the results of the WMT Shared Task on Large Scale Machine Translation Evaluation for African Languages~\cite{adelani-EtAl:2022:WMT}. In particular, we use the same system and approach as~\citet{alam-anastasopoulos:2022:WMT}, which ranked third in the Shared Task.\footnote{We note that this is the best-performing system that is publicly available -- to our knowledge, the first two performing systems were industry submissions without publicly released models or code.} These models are based on the DeltaLM~\cite{ma2021deltalm} pre-trained model, which is the adapted through fine-tuning on 24 African languages (note that Bemba is not included), as well as English and French.
The adaptation happens using adapter units~\cite{pfeiffer2020adapterhub} organized in a hierarchy following language typology~\cite{faisal-anastasopoulos-2022-phylogeny} so that similar languages share similar "family" adapters. We also compare against a baseline that finetunes the whole DeltaLM model on our training set.

Here, we only use our training data to fine-tune the adapter units for Bemba, and evaluate on both our test set as well as on the publicly available FLORES-200 devtest~\cite{nllb2022}. The results are presented in Table~\ref{tab:results-mt}, where we report sentencepiece-BLEU~\cite{nllb2022} with the FLORES-200 tokenizer. 
In general, translating into English seems to perform well, especially for the phylogeny-based model. 

The difference between the performance in the two test sets can be explained by the difference of domains. All BIG-C training data are from dialogues, while the FLORES-200 evaluation dataset is comprised of translated Wikipedia articles. Of course, larger and more diverse data collection in the future should help mitigate these issues and allow us to build general translation systems capable of handling various domains adequately.

\begin{figure*}[h!]
\centering
\tiny
\begin{tabular}{cc}
    \begin{tabular}{>{\raggedright\arraybackslash}p{.2\textwidth}>{\raggedleft\arraybackslash}p{.2\textwidth}}
    \multicolumn{2}{c}{\includegraphics[width=.35\textwidth]{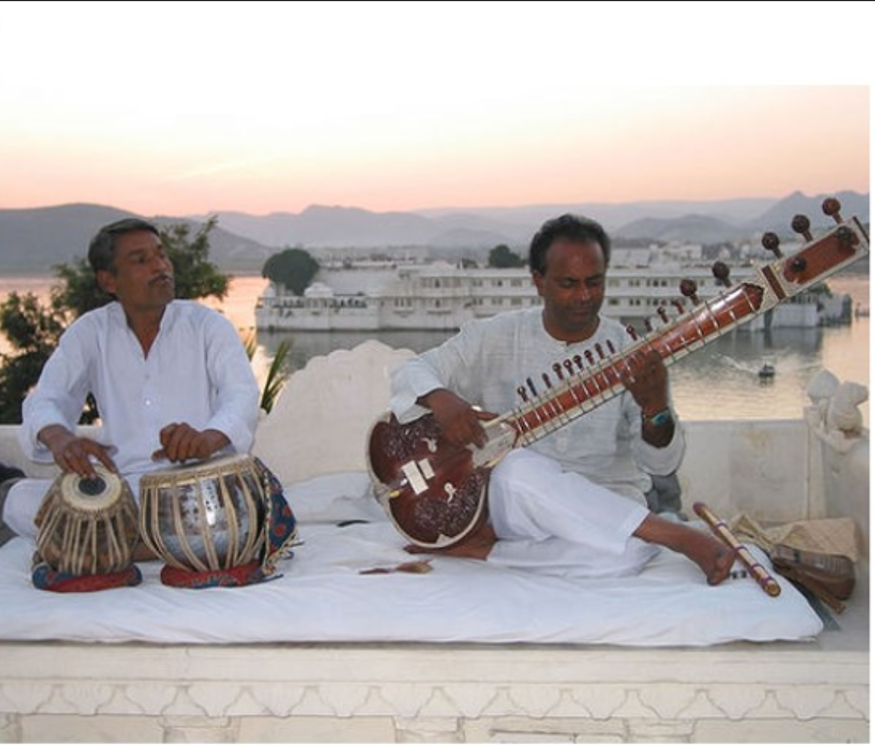}} \\
    \includegraphics[width=.6cm]{images/face.png} \includegraphics[width=2cm]{images/wavform.png} &  \\[-2em]
    & \includegraphics[width=2cm]{images/waveform2.png} \includegraphics[width=.6cm]{images/face.png} \\[-2em]
        
    Bakemba babili bamwenye bali pa mutenge mupepi ne cishiba. Balelisha banjo, ingoma na masese & \\
    \textcolor{gray}{\textit{Two Indian musicians are on a roof top near a water body. They are playing a banjo, some drums and some beads that rattle.}}& \\[-6em]
     & Aba bakemba babili ba mwenye nibashitata abafwele ifyakufwala ifya buta napantu bekele balelisisha nipa nsalu yabuta. \\
     & \textcolor{gray}{\textit{These two Indian musicians are elderly men wearing white clothes and are seated on a white cloth}}\\[-1em]
     Nalimo ukwimba kwabo kwa kupepa. icimbo ca mapepo ntile &  \\
     \textcolor{gray}{\textit{They seem to be singing religious songs. I am sure they are singing religious songs!}} & \\[-3em]
     & Emukwai. Ukwimba kwabo kulemoneka nakalimo tekwimbafye iyo, kwati nintambi. Emo basangila umutende nobutusho ngabaleimba kumipashi yabo. \\
     & \textcolor{gray}{\textit{That's right. Their singing doesn't seem to be more singing, it seems more like a religious practice. I am sure they find peace and rest as they sing to their gods}}\\[-7em]
     kwena pantu bali nipa muulu wa cikulwa nalimo beleishibisha abanabo ukutl ni nshita yakulongana kukupepa. & \\
     \textcolor{gray}{\textit{Surely, their being on top of that building seems to be a signal to the rest of their community that it's time to congregate and pray}}
    \end{tabular}
    & 
    \begin{tabular}{>{\raggedright\arraybackslash}p{.2\textwidth}>{\raggedleft\arraybackslash}p{.2\textwidth}}
    \multicolumn{2}{c}{\includegraphics[width=.35\textwidth]{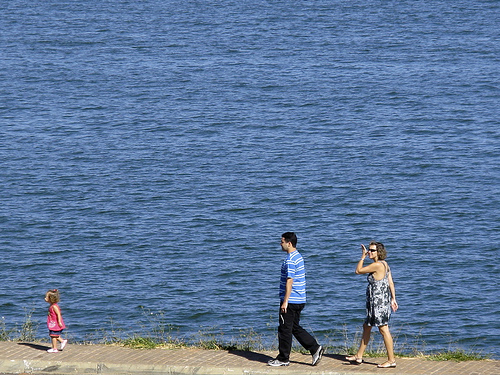}} \\
    \includegraphics[width=.6cm]{images/face.png} \includegraphics[width=2cm]{images/wavform.png} &  \\[-2em]
    & \includegraphics[width=2cm]{images/waveform2.png} \includegraphics[width=.6cm]{images/face.png} \\[-2em]
    Ba nyina , ba wishi nomwana baleenda mumbali ya cimana nabatangisha nomwana. & \\
    \textcolor{gray}{\textit{The father, wife and child walking in front of them by the riverside.}}& \\[-4em]
     & Icimana cilemoneka icikulu saana bushe aba bafyashi tabalemona ati umwana kuti aponenamo fyaleta ubwafya? \\
     & \textcolor{gray}{\textit{This river is so huge and deep, are they not afraid of the child in front to slip off and fall?}}\\[-3em]
     Awe nifyofine, cikulu icimana ici icakweba ati ngaponenamo kuti bafilwa napakutampila ukumufwaya. &  \\
     \textcolor{gray}{\textit{It is so big indeed, such that if the child fell in they would struggle so much.}} & \\[-3em]
     & Caliba icikankala saana abafyashi ukulolekesha pabana,pantu ngatabalelolekesha pa bana ngabali kuncende ngeshi kuti caleta ubwafya ubukalamba saana.\\
     & \textcolor{gray}{\textit{It is quite important for parents to ensure their children's safety, especially when outing to suchlike places because it would be a fatal encounter here.}}\\[-7em]
     Ee nifyo elo cipalile kwati umwana nasansamuka pakumutwala ku menshi, alemonekafye uwansansa & \\
     \textcolor{gray}{\textit{That is so true, and the very child is very excited to be brought to this place.}} 
    \end{tabular} 
    \\
    
    \begin{tabular}{>{\raggedright\arraybackslash}p{.2\textwidth}>{\raggedleft\arraybackslash}p{.2\textwidth}}
    \multicolumn{2}{c}{\includegraphics[width=.35\textwidth]{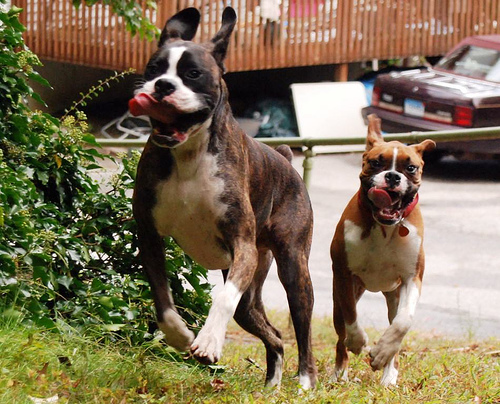}} \\
    \includegraphics[width=.6cm]{images/face.png} \includegraphics[width=2cm]{images/wavform.png} &  \\[-2em]
    & \includegraphics[width=2cm]{images/waveform2.png} \includegraphics[width=.6cm]{images/face.png} \\[-2em]
    Imbwa shibili shileingila paka panga shilebutuka. & \\
    \textcolor{gray}{\textit{Two dogs are headed to a thicket.}}& \\[-1.5em]
     & boi imbwa ishi shilemoneka ishikali,nashifumya nendimi panse kwati shamona akakulya akanona. \\
     & \textcolor{gray}{\textit{Dear these are dogs that seem to be fierce, just their race is hunty, as if after some fatty food.}}\\[-3em]
     Shifwile shileyangalafye. Imbwa shalitemwa ukubutauka, kuti wasanga limbi pali abashipepeke. &  \\
     \textcolor{gray}{\textit{These dogs must be just playing, as dogs naturally love running around.}} & \\[-2em]
     & boi ishi nimbwa shakweba ati ngawashimonafye ufwile watampako nolubilo, utunwa natukulisha.\\
     & \textcolor{gray}{\textit{No way my friend, these are dogs you run away from the moment you see them. Their mouths are too big.}}\\[-3em]
     ubwafya walifulisha umwenso, imbwa shalitemwa ukwangala nabantu,ngawabutuka ninshi wailetelelafye. & \\
     \textcolor{gray}{\textit{The problem is that you are full of cynophobia, dogs are friendly to humans and enjoy man's company.}} \\
    \end{tabular}
    &
    \begin{tabular}{>{\raggedright\arraybackslash}p{.2\textwidth}>{\raggedleft\arraybackslash}p{.2\textwidth}}
    \multicolumn{2}{c}{\includegraphics[width=.35\textwidth]{image_sample_s4.jpg}} \\
    \includegraphics[width=.6cm]{images/face.png} \includegraphics[width=2cm]{images/wavform.png} &  \\[-2em]
    & \includegraphics[width=2cm]{images/waveform2.png} \includegraphics[width=.6cm]{images/face.png} \\[-2em]
    Umulumendo aleyangala na Honda mumulu, muchibansa muli na ibumba lyabantu abaletamba uyu mulumendo. & \\
    \textcolor{gray}{\textit{A gentleman is on his motorbike spinning with a crowd of people around watching.}}& \\[-5em]
     & Abantu bonse mwibumba baletota, cilemoneka kwati nabasekalamo sana pafyo uyu muntu alepilibausha icela cakwe. \\
     & \textcolor{gray}{\textit{Everyone is excited and happy to see how he is drifting his machine.}}\\[-1em]
     Boi amangalo ya ifi yalaleta abantu abengi chapamo, balomfwa bwino ukutamba umuntu alecita ifintu ifyo abengi teti bacite. &  \\
     \textcolor{gray}{\textit{My dear this event is such a big thing, many people come by to watch and enjoy how that one can do what exceptionally.}} & \\[-5em]
     & Nomba nangu bengomfwa bwino, umunabo ngaicena akacula eka nabalupwa bakwe.\\
     & \textcolor{gray}{\textit{However the crowd when you are hurt you are on your own with relatives only.}}\\[-1em]
     Boi umuntu pakucita ifi ninshi pali cimo, ubu ubwangalo bukulu saana,limbi balapela indalama ishingi saana kuli uyo uwacimfya. & \\
     \textcolor{gray}{\textit{Dear for one to participate in anything there must be a reason, this sport is well sponsored and the winner is awarded unreservedly.}} \\
    \end{tabular}
\end{tabular}
\caption{Examples of the BIG-C dataset. The grounding image (top) and the ensuing Bemba dialog transcribed and translated in English.}
\label{fig:moreexample}
\vspace{4em}
\end{figure*}

\subsection{Other Tasks}
The authors of this study unfortunately lack the financial and compute resources, as well as required expertise, to provide baseline results for additional multimodal tasks. Nevertheless, we devote this subsection to outlining some other potential downstream uses of BIG-C.
\begin{itemize}[nolistsep,noitemsep,leftmargin=*]
    \item \textbf{Image Captioning} \ \ The dataset could be used directly for image captioning in Bemba (or English), by pairing the images with the first utterance of the conversation, which will largely function as a caption by design. 
    \item \textbf{Multimodal Language Modeling} \ \ Similarly, the corpus could be used for language and vision pre-training, and particularly to study multilingual approaches (in a field that has largely focused solely on English).
    \item \textbf{Multimodal Dialogue Modeling} Similar to other image-grounded tasks (see \S\ref{sec:related-work}), one could use to BIG-C to study dialogue, with a focus on open-ended but still grounded conversation. One could also use our dialogues as (pre-)training data for chatbots in Bemba, which could then potentially be adapted to handle specific goals or domains with fewer in-domain data.
    \item \textbf{Multimodal Translation} \ \ In the experiments above we did not take advantage of the image when translating. One could explore whether multimodal machine translation approaches~\cite[\textit{; inter alia}]{barrault-etal-2018-findings} could improve downstream performance in these resource-scarce settings.
    \item \textbf{Cross-Cultural NLP} A major limitation of our dataset (also discussed in the relevant Limitations section) is that most of the images that we use are not particularly relevant to the Zambian or sub-Saharan African context. We plan to mitigate this issue by collecting an addendum to BIG-C with images crowd-sourced \textit{in Zambia}. 
    
    Nevertheless, this limitation is simultaneously an opportunity to study cross-cultural understanding as well as the priors/assumptions/biases that speakers with a certain background exhibit. To highlight this potential, we show some additional interesting examples from BIG-C in Figure~\ref{fig:moreexample}. In the top-left example, the first speaker's utterances reveal several assumptions: that the musicians are ``Indian" (likely correct, since this image is located in India); that they ``are on a roof" (correct); that they ``sing religious songs" (unsupported); or that ``it's time to congregate and pray" (unsupported). In the example in the top-right, the first speakers assumes the image is ``by the riverside", and not e.g., by the seaside or lakeside.\footnote{Note that Zambia is a land-locked country.} 
\end{itemize}

\section{Conclusion}
\label{sec:conclusion}
In this paper, we presented a multimodal corpus comprised of multi-turn dialogues between speakers of the Zambian language, Bemba, grounded on images, transcribed and translated into English. It contains over 92,000 utterances/sentences, 180 hours of speech grounded over 16,000 images. The dataset aims to fill multiple roles: enable development of fundamental tools like speech recognition, machine translation and speech-to-text translation systems between Bemba and English; serve as a benchmark for academic and industry research; and to facilitate research in language grounding and multimodal model development towards building context-based dialogue agents, among other potential use cases. We have also provided baseline for ASR, MT and ST task. 

In future work, we plan to conduct multimodal baseline experiments, as well as attempt to mitigate the image diversity limitation by collecting an addendum to BiG-C using images taken locally in Zambia. In addition, we plan to further expand to other Zambian languages such as Tonga, Tumbuka, Chewa, or Lozi, by translating the existing dataset (creating an $n$-way parallel corpus for Zambian languages) and by direct data collection. Further down the roan we plan to study the dialectal varieties of Bemba and the other languages, by collecting contrastive datasets from different regions of the country.

\section*{Limitations}
We observe the following limitations with the dataset:
\begin{itemize}[noitemsep,nolistsep]
    \item \textbf{Language Diversity:} In terms of number of languages, the presented dataset only covers two languages; Bemba and English.
    \item \textbf{Image Diversity} All the images used in this dataset were obtained from Flickr30K image dataset. Therefore, in terms image composition, our dataset is limited to the image diversity in the Flickr30K dataset. It mostly lacks images that could be considered as "culturally relevant" ones for the Zambian or generally sub-Saharan African context. We plan to mitigate this in future work.
\end{itemize}
\section*{Ethics Statement}
We make the following declarations for the ethics statement:
\begin{itemize}[noitemsep,nolistsep]
    \item \textbf{Research:} This work was carried out mostly in Zambia, and most authors are native speakers of Bemba who also worked as validators for the data collection process. 
    \item \textbf{Participants:} All project participants; transcribers, translators and speakers/recorders were informed about the goals of the project and they signed consent forms to participate. All participants were monetarily compensated at around \$20/h for all their work.
    \item \textbf{Personal Identifiable Information:} All information that can potentially be regarded as PII such as names of participants, IDs have been removed for anonymity and will not be released with the dataset.
    \item \textbf{Copyright:} There is no potential copyright matters associated with the data contained in this dataset. We are publicly releasing the dataset under the Creative Commons BY-NC-ND 4.0 license.
\end{itemize}

\section*{Acknowledgements}
We would like to thank all the participants that were involved at different stages of the dataset creation process. We would also like to thank Desmond Elliott and Graham Neubig for insightful conversations and constructive feedback at earlier stages of our project. This project would not have been possible without generous funding by the LacunaFund. Antonios Anastasopoulos is also supported by NSF-NEH grant BCS-2109578.

\bibliography{anthology,custom}
\bibliographystyle{acl_natbib}

\clearpage

\appendix
\onecolumn
\section{Language Map of Zambia}
\label{sec:languagemap}
\begin{figure*}[h!]
    \centering
    \includegraphics[width=.8\textwidth]{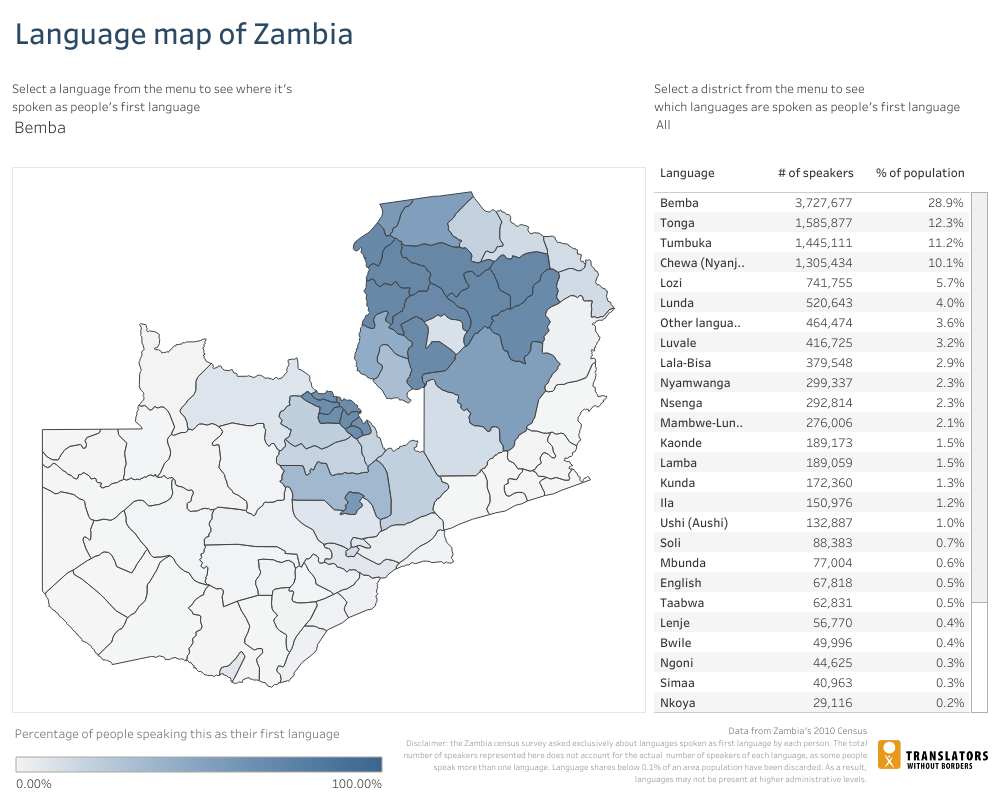}
    \caption{Language Map of Zambia. Created by Translators without Borders. Retrieved from \url{https://translatorswithoutborders.org/languages-of-zambia-interactive-en} on January 2022.}
    \label{fig:map}
\end{figure*}

\appendix
\onecolumn
\section{Participant Training Exercise}
\label{sec:training-instruction}

The following instructional steps depict the participants exercise/tutorial during a training exercise session before actual recording. The instructions were given to a pair of participant. The objective was to create a text conversations for 5 sample images in a specified image folder using Google Sheets. The recording session followed the same process, except with additional instructions involving the use of the LIG-Aikuma~\cite{Gauthier2016} app.

\begin{itemize}
    \item \textbf{STEP 1}: Open the first image in your image folders. If you are P16, for example, Go to \texttt{P1\_Session\_01 > Image7501 > Speaker\_01} [If you are \texttt{Speaker 1}] or \texttt{Speaker\_02} [If you are \texttt{Speaker 2}]. Open any of the images in the folder.
    \item \textbf{STEP 2}: While you are able to view the image, open the spreadsheet. Now that you have both image and spreadsheet opened.
    \item \textbf{STEP 3}: \texttt{Speaker 1} should enter the image number (in this case, 7501) in cell A3.
    \item \textbf{STEP 4}: \texttt{Speaker 1} should describe what is in the image by a single sentence in cell B3. The description should be a single sentence giving a clear mental picture of what is in the image. 
    \item \textbf{STEP 5} : \texttt{Speaker 2} should be able to respond to \texttt{Speaker 1} by entering their response in C3. The response can be a question, a statement or an addition to what \texttt{Speaker 1} said. As long as it's a sentence in Bemba. Remember this is a conversation and it should be able to naturally flow.
    \item \textbf{STEP 6}: \texttt{Speaker 1} should complete cell D3 with a sentence in response to what \texttt{Speaker 2} texted in cell C3.
    \item \textbf{STEP 7}: \texttt{Speaker 2} should put a response in cell E3 in response to what \texttt{Speaker 1} texted in cell D3. 
    \item \textbf{STEP 8}: \texttt{Speaker 1} closes the conversation with a sentence, however it may be in cell F3.
    \item \textbf{STEP 9}: If you have successfully generated the conversation/dialogue in the spreadsheet for the first image, then go ahead and do so for the next 4 images.
    
\end{itemize}

\end{document}